\documentclass[final]{cvpr}

\usepackage{times}
\usepackage{epsfig}
\usepackage{graphicx}
\usepackage{amsmath}
\usepackage{amssymb}
\usepackage{enumerate}
\usepackage{enumitem}

\usepackage{array}
\usepackage{verbatim}
\usepackage[pagebackref=true,breaklinks=true,colorlinks,bookmarks=false]{hyperref}

\def\cvprPaperID{4041} 
\def\confYear{CVPR 2021}

\begin{document}
	
	\setlist[enumerate, 1]{label=$\vcenter{\hbox{\large$\bullet$}}$, left=0pt, labelwidth=0.5\parindent, topsep=\parsep, parsep=0pt, itemsep=0.5\parsep}
	
	\title{CT-Net: Complementary Transfering Network for Garment Transfer with Arbitrary Geometric Changes }

	\author{Fan Yang, 
	Guosheng Lin\thanks{Guosheng Lin is the corresponding author} \\
	S-Lab, Nanyang Technological University\\
	School of Computer Science and Engineering, Nanyang Technological University \\
	{E-mail: \tt\small yyyfan3@gmail.com, \tt\small  gslin@ntu.edu.sg}}

	\maketitle
	
	\pagestyle{empty}  
	\thispagestyle{empty} 
	
	\begin{abstract}
	Garment transfer shows great potential in realistic applications with the goal of transfering outfits across different people images. However, garment transfer between images with heavy misalignments or severe occlusions still remains as a challenge. In this work, we propose Complementary Transfering Network (CT-Net) to adaptively model different levels of geometric changes and transfer outfits between different people. In specific, CT-Net consists of three modules: \romannumeral1) A complementary warping module first estimates two complementary warpings to transfer the desired clothes in different granularities. \romannumeral2) A layout prediction module is proposed to predict the target layout, which guides the preservation or generation of the body parts in the synthesized images. \romannumeral3) A dynamic fusion module adaptively combines the advantages of the complementary warpings to render the garment transfer results. 
	Extensive experiments conducted on DeepFashion dataset demonstrate that our network synthesizes high-quality garment transfer images and significantly outperforms the state-of-art methods both qualitatively and quantitatively. Our source code will be available online.
	\end{abstract}
	
	\section{Introduction}
	
\begin{figure}[ht]
		\begin{center}
			\includegraphics[width=\linewidth]{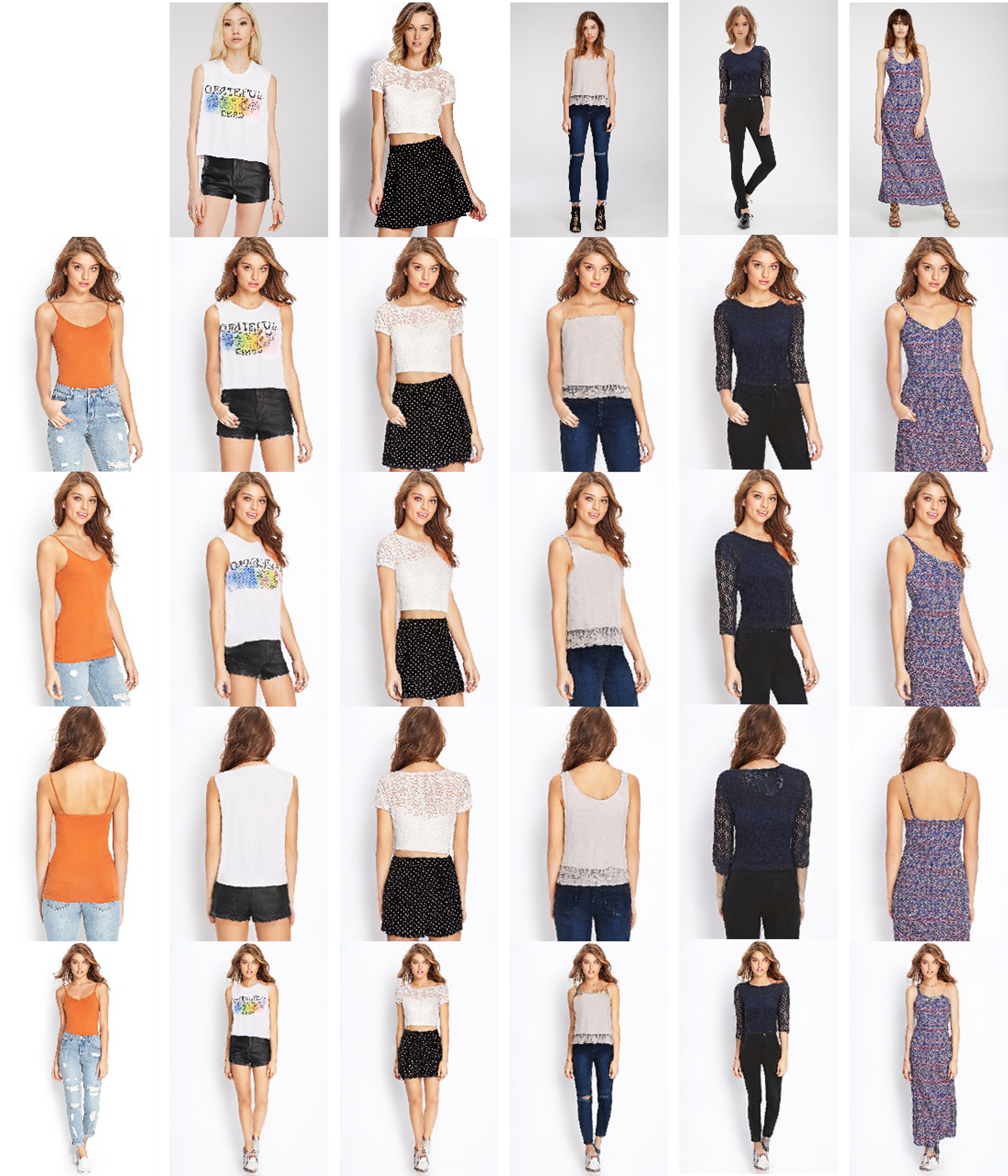}
		\end{center}
		\caption{\textbf{Garment transfer results generated by CT-Net. First row: model images. First column: target person images.} As shown above, CT-Net naturally transfers clothes across different people with arbitrary poses or shapes and synthesizes photo-realistic images with well-preserved characteristics of the desired clothes and distinct identities of humans. Please refer to \textit{supplementary materials} for more results.}
		\label{fig:R1}
	\end{figure}

	Most existing virtual try-on methods are based on simplifying assumptions: (\romannumeral1) Pure clothing images or 3D information are available. (\romannumeral2) Pose changes are simple without heavy misalignments or severe occlusions. We argue that these simplifying assumptions greatly limit the application scope of these methods in the realistic virtual try-on scenarios. To address this issue, we propose Complementary Transfering Network (CT-Net), a novel image-based garment transfer network that does not rely on pure clothing images or 3D information while capable to adaptively deal with different levels of geometric changes. As shown in Figure~\ref{fig:R1}, given a target person image $I^T$ and a model image $I^M$, without any restriction to the poses or shapes of $I^T$ and $I^M$,  our CT-Net synthesizes photo-realistic garment transfer results, in which the person in $I^T$ wearing the clothes depicted in $I^M$ with well-preserved details. 
	
	Despite various methods have been proposed to realize virtual try-on in different settings~\cite{han2018viton, wang2018toward, yu2019vtnfp, yang2020towards, raj2018swapnet, dong2019fw, han2019clothflow, wu2019m2e}, there is still a gap between these methods and the unlimited realistic scenarios. Some methods~\cite{guan2012drape, chen2016synthesizing, pons2017clothcap} involve 3D information to deal with occlusions, but they are greatly limited by expensive devices and high computational costs. Others~\cite{han2018viton, wang2018toward, yu2019vtnfp, yang2020towards} may rely on stand-alone clothing images, which are not easy to get timely online. Moreover, most of them attempt to model the geometric changes of the clothes utilizing a Thin Plate Spline (TPS) warping. Because TPS warping is limited by a small number of parameters and only capable to shape simple deformations, their methods fail to deal with complex cases with heavy misalignments or severe occlusions. Garment transfer methods aim to transfer outfits across different people. Although prior arts~\cite{raj2018swapnet, han2019clothflow, wu2019m2e} have achieved considerable progress, none of them address the issue of large geometric changes. 
	
\begin{figure*}[t]
		\begin{center}
			\includegraphics[width=\linewidth]{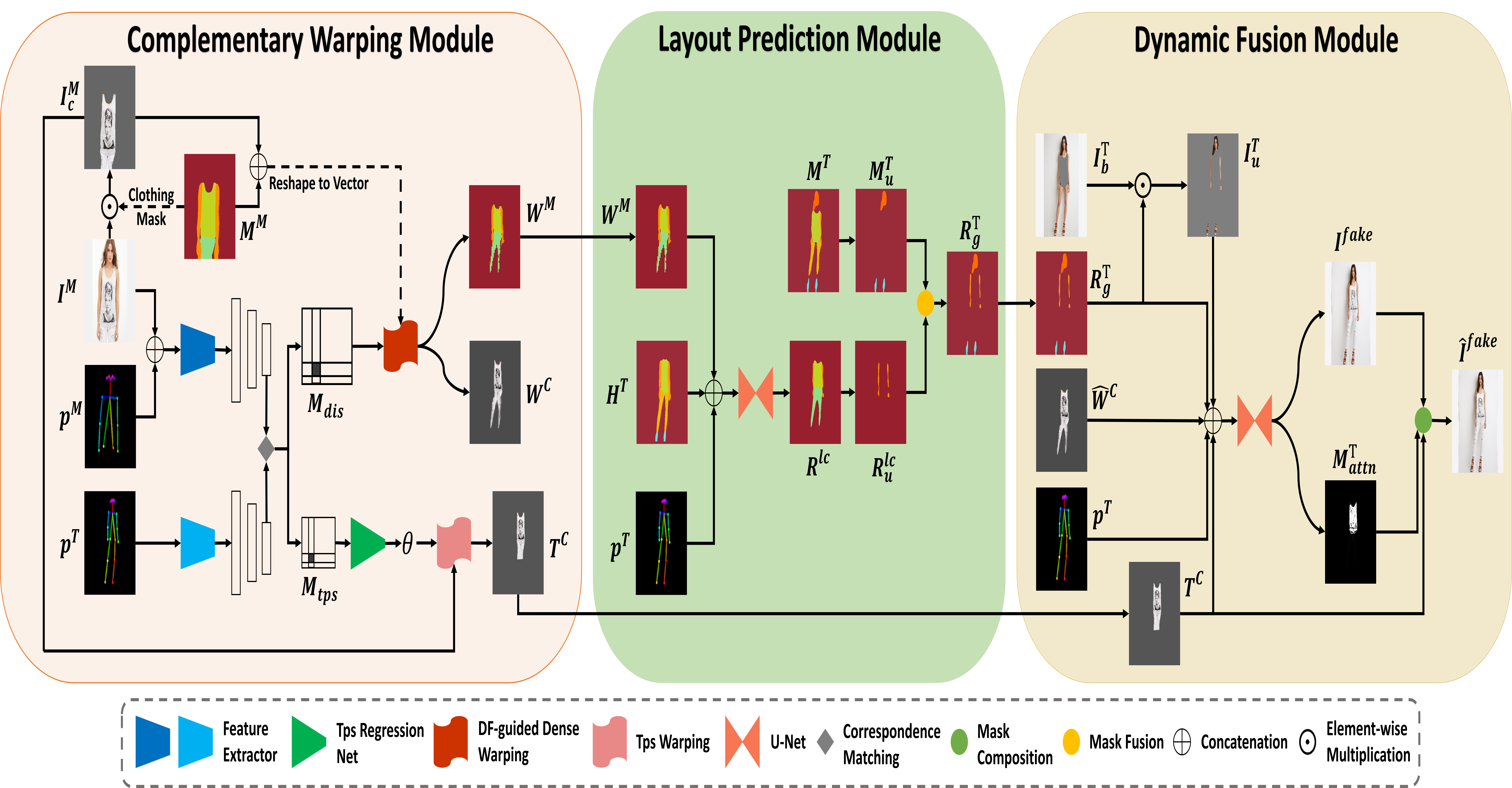}
		\end{center}
		\caption{The overall architecture of CT-Net: (\romannumeral1) Complementary Warping Module (CWM) estimates two complementary warpings to warp the desired clothes $I^M_c$ in two different granularities, where $W^{\{\cdot\}}$ denotes the warping result of DF-guided dense warping and $T^{\{\cdot\}}$ denotes the warping result of TPS warping. (\romannumeral2) Layout Prediction Module (LPM) predicts the target layout $R^T_g$ to guide the layout adaption, where $H^T$ denotes the clothing-agnostic human representations. (\romannumeral3) Dynamic Fusion Module (DFM) adaptively integrates all the information with an attention mechanism $M^T_{attn}$ to render the photo-realistic garment transfer result $\hat{I}^{fake}$. Note $\widehat{W}^C = W^C \odot R^{lc}_{c}$, where $R^{lc}_{c}$ denotes the predicted clothing mask in $R^{lc}$, which is not shown in the figure for simplicity. }
		\label{fig:pipeline}
	\end{figure*}

	We aim to fulfill this gap by proposing a novel garment transfer network, Complementary Transfering Network (CT-Net), which precisely transfers outfits across different people while tolerating different levels of geometric changes. As shown in Figure~\ref{fig:pipeline}, CT-Net has three modules:
	
	First, a Complementary Warping Module (CWM) is introduced to warp the desired clothes into the target region. Specially, we simultaneously estimate two complementary warpings with different levels of freedom: (a) Distance fields guided (DF-guided) dense warping. (b) Thin Plate Spline (TPS) warping. DF-guided dense warping has a high degree of freedom and is utilized to warp the desired clothes to be well-aligned with the target pose; while limited by a small number of parameters, TPS warping roughly transfers the desired clothes into the target region with well-preserved textures. 
	
	Second, a Layout Prediction Module (LPM) is introduced to predict the target layout, in which the target person wearing the desired clothes. Compared to prior works, which may suffer from the misalignments between inputs~\cite{raj2018swapnet, han2019clothflow, yang2020towards}, our Layout Prediction Module makes more accurate predictions based on the aligned warping results from Complementary Warping Module. Leveraging the predicted target layouts, our network dynamically determines the non-target body areas and the occluded body areas, which guides the adaptive preservation and generation. Benefited from joint training, Layout Prediction Module also adds spatial constraints to the training of complementary warpings, encouraging the warping results to be more coherent with the target person. 
	
	Third, a Dynamic Fusion Module (DFM) integrates all the information provided by previous modules to render the garment transfer results. Specifically, our Dynamic Fusion Module adopts an attention mechanism to adaptively combine the advantages of the two complementary warpings and synthesizes photo-realistic garment transfer results with well-preserved characteristics of the clothes.
	
	Extensive experiments conducted on DeepFashion dataset demonstrate the superiority of our method compared to the state-of-art methods. Our main contributions can be summarized as follows:
	\begin{enumerate}
		\item We propose a novel image-based garment transfer network, which adaptively combines two complementary warpings to model different levels of geometric changes and synthesizes photo-realistic garment transfer results with well-preserved characteristics of the clothes and distinct human identities. 
		\item A novel Layout Prediction Module makes precise prediction on the target layout, which clearly shapes the synthesized results, guides the adaptive preservation and generation of the body parts and adds spatial constraints to the training of the complementary warpings.
		\item Evaluated on DeepFashion~\cite{liu2016deepfashion} dataset, CT-Net synthesizes high-quality garment transfer results and outperforms all the state-of-art methods both  qualitatively and quantitatively.
	\end{enumerate}
	
	\section{Related Work}
	\textbf{Generative Adversarial Networks.}
	Generative Adversarial Networks (GANs)~\cite{goodfellow2014generative} have been demonstrated very effective in generating fake images, which are indistinguishable from the real ones in the original dataset. Conditional GAN (cGAN)~\cite{mirza2014conditional} adopts extra information to further control the generation results, which promotes the development of many applications~\cite{hsiao2019fashion++,wang2018high,park2019semantic,albahar2019guided,han2019finet}. Specifically, Isola \etal~\cite{isola2017image} proposed an image-to-image translation network to transfer images from one domain to another, which explores relationships across different domains. Similarly, we also employ a cGAN to synthesize photo-realistic garment transfer results conditioned on the desired clothes and the target pose.
	
	\textbf{Pose-guided Human Image Generation.}
	Ma \etal~\cite{ma2017pose} made an early attempt to generate human images conditioned on pose with a two-stage network. Esser \etal~\cite{esser2018variational} proposed a conditional U-Net to disentangle the pose and appearance. More recent methods~\cite{siarohin2018deformable, dong2018soft, ren2020deep, han2019clothflow} propose to solve this problem utilizing warp-based methods.
	Zhu \etal~\cite{zhu2019progressive} employs a sequence of pose-attentional transfer
	blocks to progressively deal with large pose discrepancies.
	Zhang \etal~\cite{zhang2020cross} for the first time introduces cross-domain semantic matching, which learns dense correspondence warping between cross-domain inputs. Inspired by \cite{zhang2020cross}, we also employ the dense correspondence warping. However, we focus on the exact problem of garment transfer and estimate a distance fields guided dense warping. Benefited from the joint training of all the modules, our distance fields guided dense warping naturally warps the clothing items to be well-aligned with the target pose and preserves the clothing patterns well.

	\textbf{Virtual Try-on.}
	Many conventional virtual try-on methods rely on 3D information~\cite{guan2012drape, chen2016synthesizing, pons2017clothcap}. Along with the advances of deep neural networks, more recent works attempt to synthesize try-on results based on 2D images.
	Various methods~\cite{han2018viton,wang2018toward,yu2019vtnfp,yang2020towards,han2019clothflow} have been proposed to transfer clothes in a stand-alone clothing image onto a target person. However, all of them rely on simplifying assumptions that the clean clothing images are available and geometric changes are simple enough to be modeled by a Thin Plate Spline warping (TPS) with a small number of parameters (\eg 6 for affine and $2\times5\times5$ for TPS as in~\cite{wang2018toward}). These assumptions greatly hinder the application of these methods in the realistic virtual try-on scenarios. Wu \etal~\cite{wu2019m2e} proposed 
	to use densepose~\cite{alp2018densepose} descriptor to warp the desired clothes onto the target person. But warping estimated by densepose descriptor can be very sparse when there are large occlusions, leading to unconvincing synthesized results. Methods mentioned above only focus on the transfer of upper clothes. SwapNet~\cite{raj2018swapnet} employs a two-stage network to transfer the entire outfits across people images. To deal with the misalignments of features, they adopt ROI pooling and encode each clothing regions into high-dimensional features. 
	However, the encoded features are inadequate to preserve the local textures, which leads to blurry synthesized results. Different from these methods, we explore a wilder application scope by adaptively combining two complementary warpings to model different levels of deformations and synthesizing high-quality garment transfer results with arbitrary geometric changes.

	\section{Complementary Transfering Network}
	Given the image $I^M$ depicting the model wearing desired clothes, the image $I^T$ depicting the target person, assuming clothes, poses and shapes of $I^M$ and $I^T$ can be arbitrary, our goal is to synthesize high-quality garment transfer results with well-preserved characteristics of the clothes and distinct human identities.
	To achieve our goal, we present Complementary Transfering Network (CT-Net). As shown in Figure~\ref{fig:pipeline}, CT-Net consists of three modules. First, we introduce a Complementary Warping Module (CWM) to simultaneously estimate two complementary warpings to deal with different levels of geometric changes (Section~\ref{sec:CWM}). Second, we introduce a Layout Prediction Module (LPM), in which we predict the target layouts to guide the preservation or generation of body parts in the synthesized results (Section~\ref{sec:LPM}). The third module is a Dynamic Fusion Module (DFM), which adaptively combines the advantages of the complementary warpings to render the garment transfer results (Section~\ref{sec:FM}).

	\subsection{Complementary Warping Module} \label{sec:CWM}
	To synthesize garment transfer results, one of the main challenges is to combine the clothes of the model with the misaligned target pose. A good practice to address this issue is to estimate warpings between the clothes and the target pose ~\cite{siarohin2018deformable,han2019clothflow,ren2020deep,li2019dense}. However, to our best knowledge, there is no perfect warping that can shape any geometric changes. Warpings with a high degree of freedom are capable to shape large geometric changes, but they have higher error rates and may fail to preserve complex visual patterns; warping methods with limited numbers of parameters can retain the textures of the desired clothes well, but they can not deal with large geometric changes. Therefore, we propose Complementary Warping Module to simultaneously estimate two complementary warpings with different degrees of freedom to warp the inputs in two granularities, which enables our network to deal with different levels of geometric changes.
	
	As shown in Figure~\ref{fig:pipeline}, we first employ two separate feature extractors to extract high-level features. Then we match the features to calculate the correspondence matrixs, $\mathcal{M}_{dis}$ and $\mathcal{M}_{tps}$, which are then used to estimate DF-guided dense warping and TPS warping. 
	Given a model image $I^M$, we estimate the original layout $M^M$ and extract the corresponding clothes $I^M_{c}$. 
	Denote the warping results of DF-guided dense warping and TPS warping as $W^{\{\cdot\}}$ and $T^{\{\cdot\}}$.   DF-guided dense warping is utilized to transfer $I^M_{c}$ and $M^M$ in a finer granularity to get $W^C$ and $W^M$. TPS warping transfers $I^M_{c}$ to $T^C$, which is roughly aligned with the target pose.
	
	\textbf{Pose Representations.}
	We adopt the keypoint distance fields as our pose representations. In detail, we apply the state-of-art pose estimator~\cite{cao2017realtime} to estimate keypoint confidence maps of the target person and the model image. Then we convert the sparse joint maps into keypoint distance fields by replacing each zero pixel with its distance to the joint mask. Keypoint distance filed represents each pixel with a unique distance vector, which greatly facilitates the estimation of the correspondence matrixs. In this paper, pose representations of the target person and the model are denoted as $p^T$ and $p^M$.
	
	\textbf{Correspondence Matrix.}
	We adopt the keypoint distance fields to estimate dense correspondence matrixs. To be specific, let $\mathcal{F}_A$, $\mathcal{F}_B$ denote the separate feature extractors, we first extract high-level features $m_f \in \mathbb{R}^{H \times W \times C}$ and $t_f \in \mathbb{R}^{H \times W \times C}$ as follows:
	\begin{gather}
		m_f = \mathcal{F}_A(I^M, p^M), \\
		t_f = \mathcal{F}_B(p^T),
	\end{gather}
	where $I^M$ denotes the model image.
	
\begin{figure}[ht]
		\begin{center}
			\includegraphics[width=0.8\linewidth]{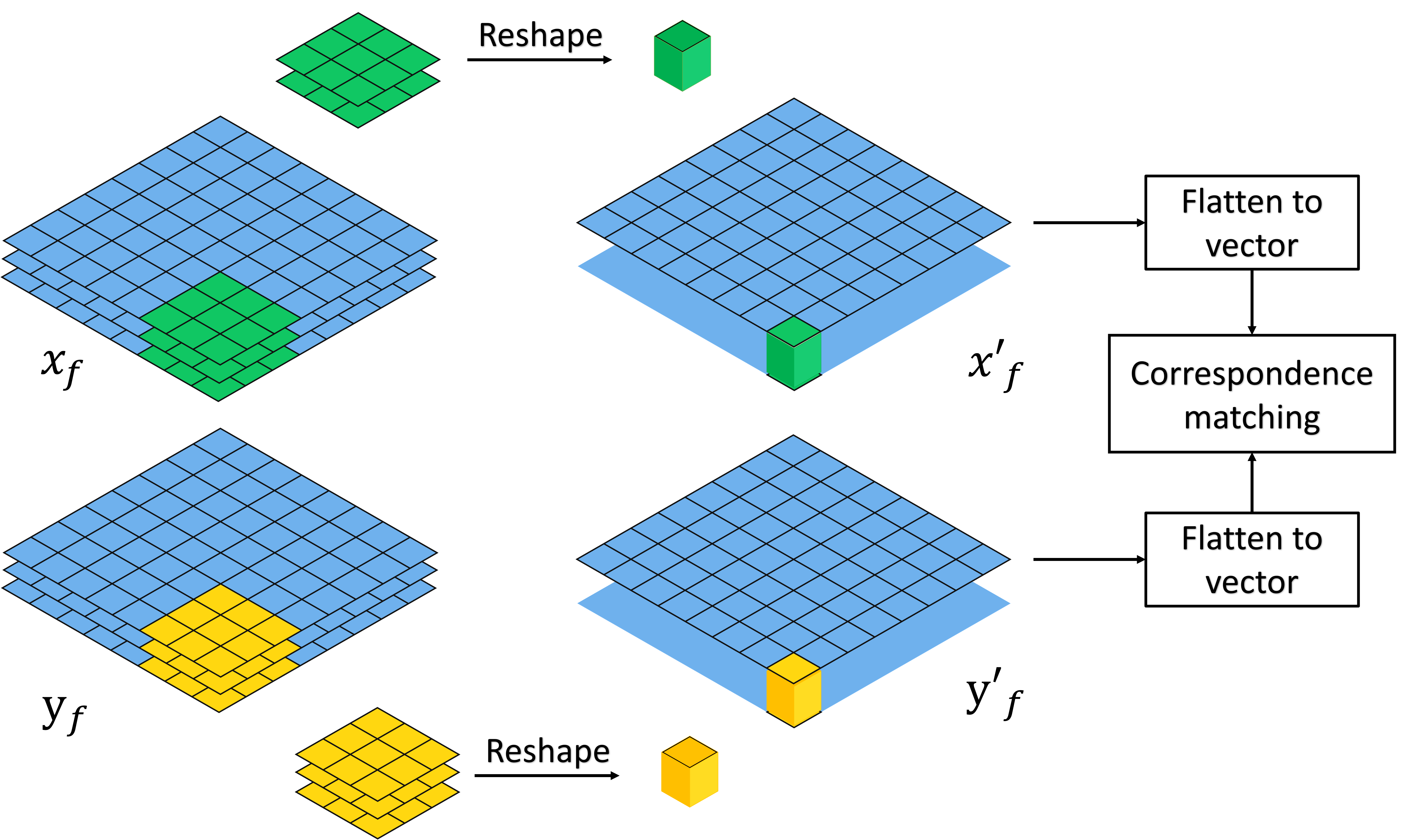}
		\end{center}
		\caption{Illustration of the correspondence matching process with sliding window of size 3, stride 1 and padding size 1. Sliding blocks are first extracted from high-level features $\{x_f, y_f\} \in \mathbb{R}^{H \times W \times C}$ and then flattened into a column of the aggregated features $\{x'_f, y'_f\} \in \mathbb{R}^{H \times W \times 9C}$, which are utilized to estimate the correspondence matrix $\mathcal{M} \in \mathbb{R}^{HW \times HW}$.
	}
		\label{fig:unfold}
	\end{figure}

	To estimate the correspondence matrixs, we aggregate the features into different scales with different sliding windows,  which is illustrated in Figure~\ref{fig:unfold}. 
	Specifically, we use sliding window of size 3, with stride 1 and padding size 1 to estimate the correspondence matrix $\mathcal{M}_{dis} \in \mathbb{R}^{HW \times HW}$ for DF-guided dense warping and we apply a sliding window of size 4, with stride 4 and padding size 0 to estimate $\mathcal{M}_{tps} \in \mathbb{R}^{HW/16 \times HW/16}$ for TPS warping. 
	
	We employ the same correspondence layer as~\cite{zhang2020cross} to match the aggregated features, which can be formulated as:
	\begin{equation}
		\mathcal{M}(i,j) = \frac{(m'_f(i)^T - u_m)  (t'_f(j) - u_t)} {\Vert m'_f(i) - u_m \Vert \Vert t'_f(j) - u_t \Vert},
	\end{equation}

	where $m'_f$ and $t'_f$ represent the aggregated features, $u_m$ and $u_t$ represent the mean vectors.
	
	
	\textbf{Distance Fields guided (DF-guided) Dense Warping.}
	According to the dense correspondence matrix $\mathcal{M}_{dis}$, We calculate the weighted average to estimate the distance fields guided (DF-guided) dense warping :
	\begin{equation}
		\mathcal{W}^X(u) = \sum_v \mathop{softmax} \limits_{v} (\alpha \mathcal{M}_{dis}(u,v)) \cdot X(v),
	\end{equation}
	where $\alpha$ is a hyper-parameter controlling the sharpness of the softmax. We set it as 100 here. DF-guided dense warping learns a dense mapping between two images with a high degree of freedom, which is capable to handle large geometric changes. We utilize it to transfer the clothes of the model image to be well-aligned with the target person, which provides important guides for the generator to reconstruct the local textures of the clothes in the synthesized results.  
	
	\textbf{TPS Warping.}
	Given the model clothes $I^M_c$, we warp it with the deformation shaped by TPS to be roughly aligned with the target person $I^T$.
	
	We estimate the TPS warping from $\mathcal{M}_{tps}$. As shown in Figure~\ref{fig:pipeline}, 
	we first employ a regression net to predict the corresponding control points and then calculate the parameters $\theta$. For training, we adopt the second-order constraint~\cite{yang2020towards} to restrict the TPS warping from generating unnatural deformations or mess textures, which is denoted as $\mathcal{L}_{sc}$.
	The total loss can be formulated as:
	\begin{equation}
		\mathcal{L}_{tps} = \lambda_1 \Vert I^T_{c} - T^C \Vert_1 + \lambda_2\mathcal{L}_{sc},
	\end{equation}    
	where $I^T_{c}$ represents the ground-truth clothes extracted from the model image, $\lambda_1$ and $\lambda_2$ are the weights for the two loss terms. Both of them are set to 10, respectively, in our experiments.
	
	\subsection{Layout Prediction Module} \label{sec:LPM}
	We propose Layout Prediction Module to predict target layouts, in which the target person wearing the desired clothes depicted in the model image. Prior works~\cite{raj2018swapnet,dong2018soft,han2019clothflow} mostly generate the target layouts conditioned on the original layouts of the model and the target pose. However, suffered from the limited receptive fields of CNNs, these methods fail to understand the correlation between the original layout and the pose representations while facing with large geometric changes. Compared to these methods, we explore a warping-based strategy to eliminate the misalignments and facilitates the prediction. 
	
	As shown in Figure~\ref{fig:pipeline}, we first align the original layout with the target pose leveraging the DF-guided dense warping and then feed the warped layout $W^M$ with the clothing-agnostic representations $H^T$ into the U-net~\cite{ronneberger2015u} to predict the target limb and clothing mask $R^{lc}$. Denote the head and shoes mask of the target person as $M^T_u$, the limb mask of the predicted layout as $R^{lc}_u$, we merge $M^T_u$ with $R^{lc}_u$ to form the complete target layouts $R^T_g$, which provides important guides for the generator to adaptively determine the preservation or generation of the body parts in the synthesized results (Section~\ref{sec:FM}).
	
	We get the original layouts utilizing the state-of-art human parsing network~\cite{liang2018look} and adopt the segmentation of densepose descriptor~\cite{alp2018densepose} as our extra clothing-agnostic representations $H^T$. Since head and shoes are not in the transfer list of our model, we remove these areas in $H^T$ and the model's layout $M^M$. For training this module, we adopt the pixel-level cross-entropy loss, denoted as $\mathcal{L}_{layout}$. Benefited from joint training, Layout Prediction Module adds extra spatial constraints to the training of Complementary Warping Module, encouraging the warping results to be more coherent with the target person.
	
	\subsection{Dynamic Fusion Module} \label{sec:FM}
	Dynamic Fusion Module is proposed to adaptively combine the advantages of the two complementary warpings to render the garment transfer results with well-preserved characteristics of the clothes and distinct identities of humans. As shown in Figure~\ref{fig:pipeline}, we first utilize the merged target layout $R^T_g$ to extract the non-target body parts from the non-clothing model image $I^T_b$. Leveraging the non-target body parts and the target layout, the generator learns to preserve the details in the non-target body parts and inpaint occluded body parts according to the target layouts, leading to well-preserved human identities and clear body boundaries in the synthesized results. We then adopt a cGAN to integrate the non-target body parts $I^T_u$, the target layout $R^T_g$, complementary warping results $\widehat{W}^C$, $T^C$ and target pose representation $p^T$ to render the initial generation result $I^{fake}$. Note $\widehat{W}^C = W^C \odot R^{lc}_{c}$, where $R^{lc}_{c}$ denotes the predicted clothing mask in $R^{lc}$, which is not shown in the Figure~\ref{fig:pipeline} for simplicity.
	
	An attention mechanism is employed to combine the advantages of the two complementary warpings, where an attention mask $M^T_{attn}$ is estimated to compose the initial generation result $I^{fake}$ with the warping result from TPS as our final garment transfer result $\hat{I}^{fake}$:
	\begin{equation}
		\hat{I}^{fake} = T^C \odot M^T_{attn} + I^{fake} \odot (1 - M^T_{attn}).  
	\end{equation}

\begin{figure}
	\footnotesize
	\setlength{\tabcolsep}{0.002\textwidth}
	\begin{center}
		\begin{tabular}{*{6}{m{0.075\textwidth}<{\centering}}}
			\includegraphics[width=0.075\textwidth]{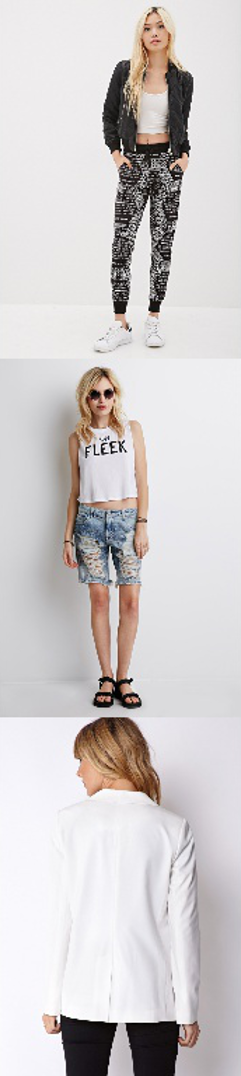} & \includegraphics[width=0.075\textwidth]{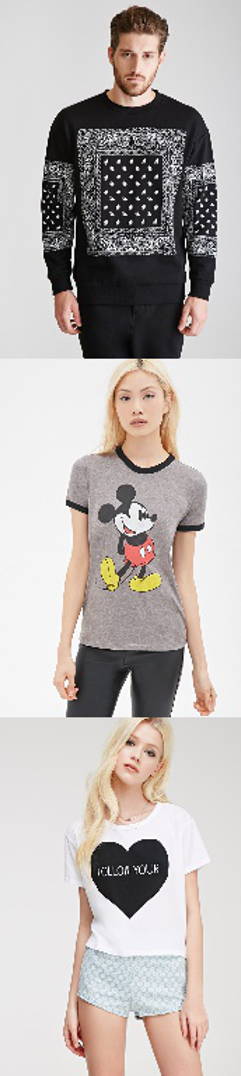} &
			\includegraphics[width=0.075\textwidth]{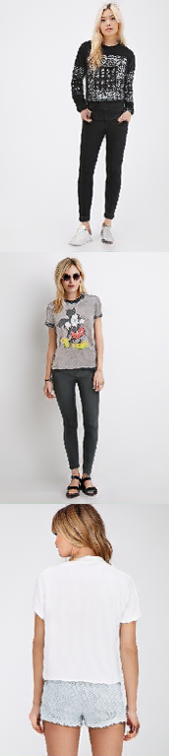} & 
			\includegraphics[width=0.075\textwidth]{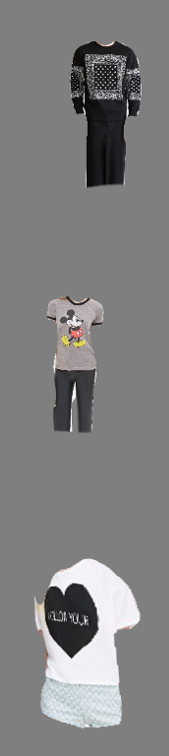} &
			\includegraphics[width=0.075\textwidth]{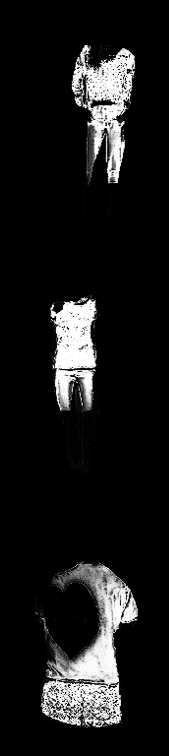} &
			\includegraphics[width=0.075\textwidth]{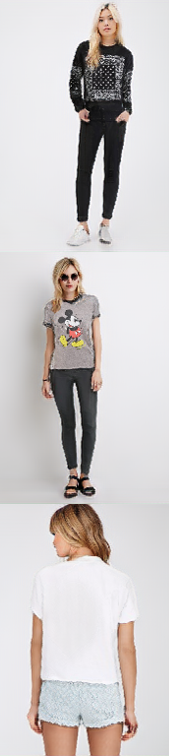} \\
			Target Person & Model Image & Initial Generation Result & TPS Warping Result & Attention Mask & Final Result \\
		\end{tabular}
	\end{center}
	\caption{Examples of our attention mechanism. From left to right: target person $I^T$, model image $I^M$, initial generation result $I^{fake}$, TPS warping result $T^C$, attention mask $M^T_{attn}$, final result $\hat{I}^{fake}$.}
	\label{fig:attn}
	\end{figure}

	As shown in Figure~\ref{fig:attn}, our attention mechanism adaptively selects different regions from the initial generation result $I^{fake}$ and the warping result of TPS according to different levels of geometric changes. For example, when the geometric change is simple and can be shaped by the TPS warping, as the first two rows in Figure~\ref{fig:attn}, the attention mechanism selects more regions on the warping result from TPS to refine the initial generation result; while as the third row, when there are heavy misalignments or severe occlusions, the attention mechanism tends to retain the initial generation result and ignore the large logos or unreasonable textures in the warping result of TPS. In this way, we adaptively combines the two complementary warpings to deal with different levels of geometric changes and expand the application scope of our model to wilder scenarios. 
	
	\subsection{Loss Functions}
	To encourage the training of different modules benefit each other, we train our model in a joint style. We combine several different losses to produce high-quality garment transfer results:
	
	\textbf{Perceptual Loss.}
	Based on the difference between high-level features, perceptual loss has been proved efficient in the image generation tasks~\cite{johnson2016perceptual}. To pose perceptual constraints on the synthesized results, we adopt a pre-trained VGG network~\cite{simonyan2014very} to extract multi-level features $\phi_{j}$ and compute the perceptual loss as:
	\begin{equation}
		\mathcal{L}_{perceptual} = \sum^N_{j=1} \lambda_j\Vert \phi_j(\hat{x}_t) - \phi_j(x) \Vert_2. 
	\end{equation}
	
	\textbf{Style Loss.}
	We further apply the style loss~\cite{gatys2015texture} to penalize the statistic error between high-level features, which can be formulated as:
	\begin{equation}
		\mathcal{L}_{style} = \sum^N_{j=1} \Vert G^{\phi}_j(\hat{x}_t) - G^{\phi}_j(x_t) \Vert_2,
	\end{equation}
	where $G^{\phi}_j$ denotes the Gram matrix estimated from $\phi_j$.
	
	\textbf{Contextual Loss.}
	To encourage our network to preserve more details from the desired clothes $I^M_c$, we employ the contextual loss proposed in~\cite{mechrez2018contextual}, which can be formulated as:
	\begin{equation}
		\begin{split}
		 & \mathcal{L}_{contextual} = \\ 
		 & \sum_{l=1} \lambda_l \left[-\log \left(\frac{1}{n_l} \sum_i \max \limits_{j} A^l (\phi^l_i(\hat{x}_B), \phi^l_j(y_B))\right)\right],
		\end{split}
	\end{equation} 
	where $A^l$ denotes the pairwise affinities between features.
	
	\textbf{Adversarial Loss.}
	To force the generator to learn the real distributions of the dataset and generate realistic human images, we deploy a discriminator to discriminate the generated fake images from the real samples in the dataset. The loss can be formulated as:
	\begin{equation}
		 \mathcal{L}_{adv} = \mathbb{E}_{x,y}[\log(\mathcal{D}(x,y))]
		 +\mathbb{E}_{x}[\log(1-\mathcal{D}(x,\mathcal{G}(x)))],
	\end{equation}
	where $x$ represents the inputs and $y$ is the ground-truth.
	
	\textbf{Objective Function.} Besides the losses above, we apply a L1 regularization $\mathcal{L}_{reg} = \Vert 1 - M \Vert_1$ on $M^T_{attn}$ to prevent the network from overfitting to the initial synthesized result $I^{fake}$. We also take a L1 loss to stabilize our training process, which can be defined as $\mathcal{L}_{l1} = \Vert \hat{x} - x \Vert_1$. Our objective function is a weighted sum of above terms:
	\begin{equation}
		\begin{split}
		& \mathcal{L}_{total} = \alpha_1 \mathcal{L}_{l1} + \alpha_2 \mathcal{L}_{tps} + \alpha_3 
		\mathcal{L}_{layout} + \alpha_4 \mathcal{L}_{perceptual} + 
		\\ &\alpha_5 \mathcal{L}_{style} + \alpha_6 \mathcal{L}_{contextual} + \alpha_7 \mathcal{L}_{adv} + \alpha_8 \mathcal{L}_{reg},
		\end{split}
		\label{total_loss}
 		\end{equation} 
	where $\alpha_i$,$(i=1,\ldots,8)$ are hyper-parameters controlling the weights of each loss.
	\section{Experiments}
	\subsection{Dataset}
	We evaluate our model on the In-shop Clothes Retrieval Benchmark of DeepFashion dataset~\cite{liu2016deepfashion}, which contains 52,712 fashion images of resolution 256 $\times$ 256. For training, we select 37,836 pairs of images depicting the same person wearing the same outfit with different poses. At test stage, we select 4,932 pairs of images which are not overlapped with the training set. As the realistic virtual try-on scenarios, each testing pair contains two different people with different clothes and poses.
	
	\subsection{Implementation Details}

	We adopt Adam~\cite{kingma2014adam} with $\beta_1=0.5, \beta_2=0.999$ as the optimizer in our all experiments. Our model is trained in stages. Complementary Warping Module is firsted trained for 20 epoches to estimate reasonable warpings. Then our model is jointly trained in an end-to-end manner for another 80 epoches. 
	Learning rate is fixed at 0.0002 for the first 40 epoches and then decays to zero linearly in the remaining steps. InstanceNorm2d Normalization~\cite{ulyanov2016instance} is applied to all layers to stabilize the training. The detailed network structures can be found in the \textit{supplementary materials}. To balance the scales of losses in Eqn.~\ref{total_loss}, we set $\alpha_{1,2,3,7,8}=10$ and $\alpha_{4,5,6}=1$.

	\begin{figure*}
	\scriptsize   
	\setlength{\tabcolsep}{0.002\linewidth}
	\begin{center}
		\begin{tabular}{*{5}{m{0.094\linewidth}<{\centering}}m{0.016\linewidth}*{5}{m{0.094\linewidth}<{\centering}}}
			Target Person & Model Image & ACGPN & CoCosNet & CT-Net(ours) & &  Target Person & Model Image & ACGPN & CoCosNet & CT-Net(ours)\\
			\includegraphics[width=\linewidth]{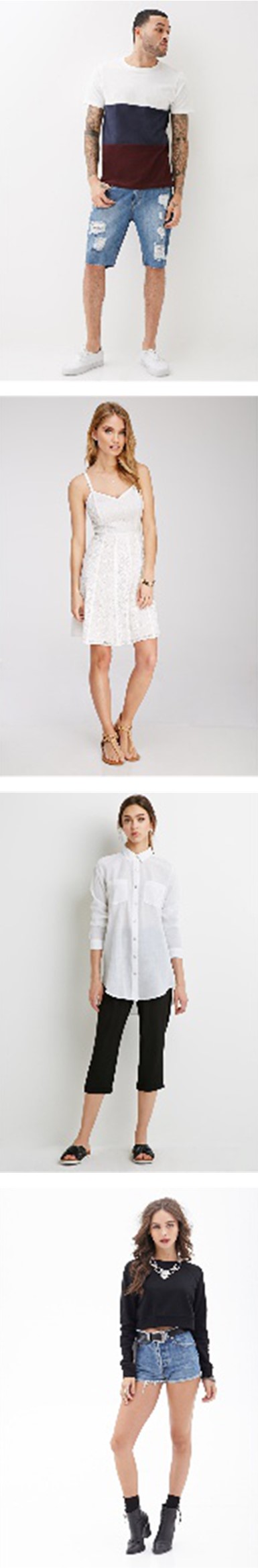} & \includegraphics[width=\linewidth]{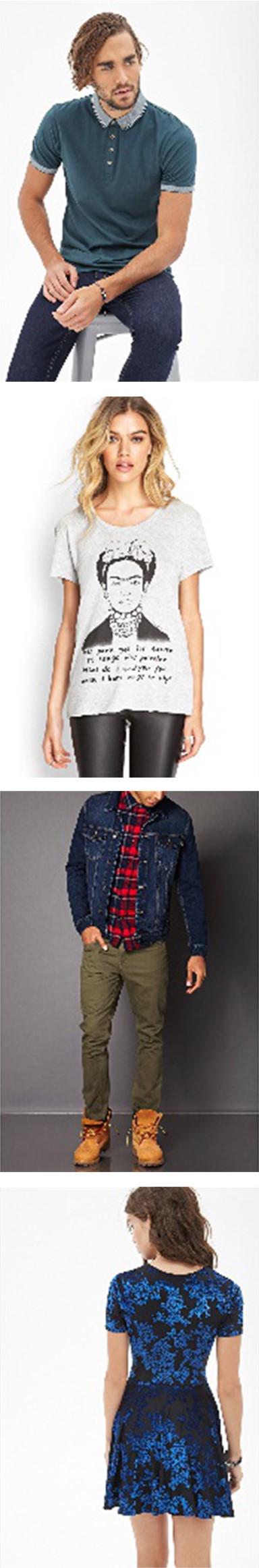} &
			\includegraphics[width=\linewidth]{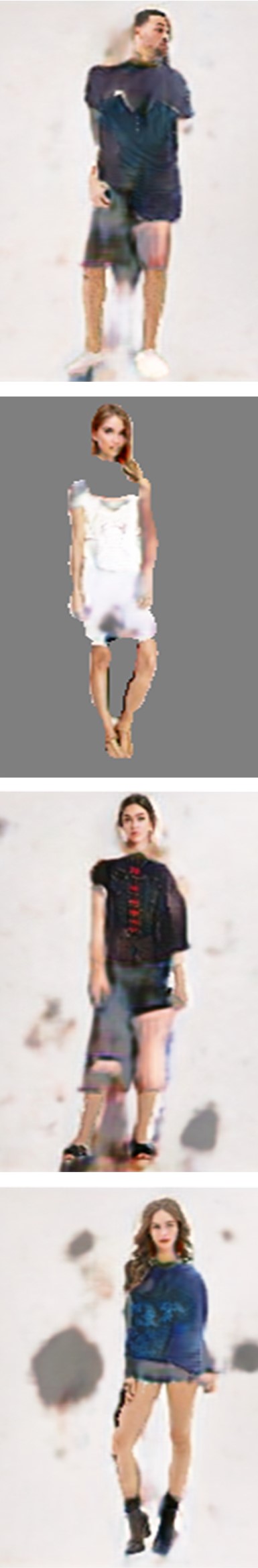} & 
			\includegraphics[width=\linewidth]{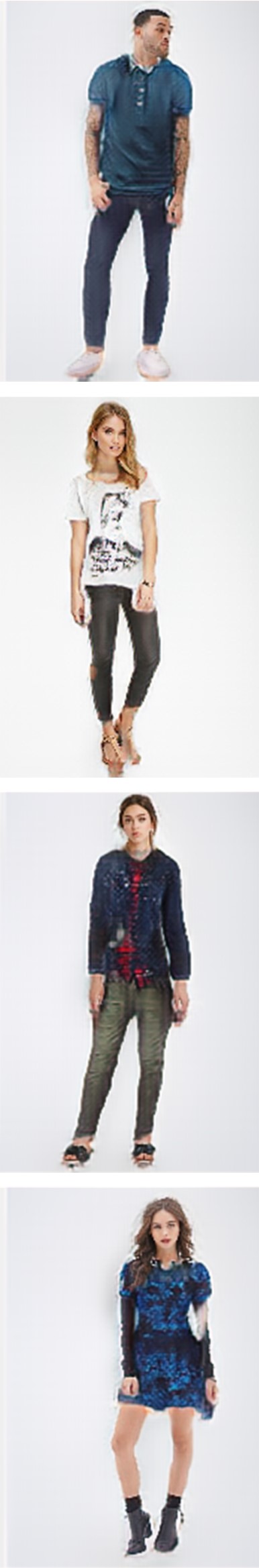} &
			\includegraphics[width=\linewidth]{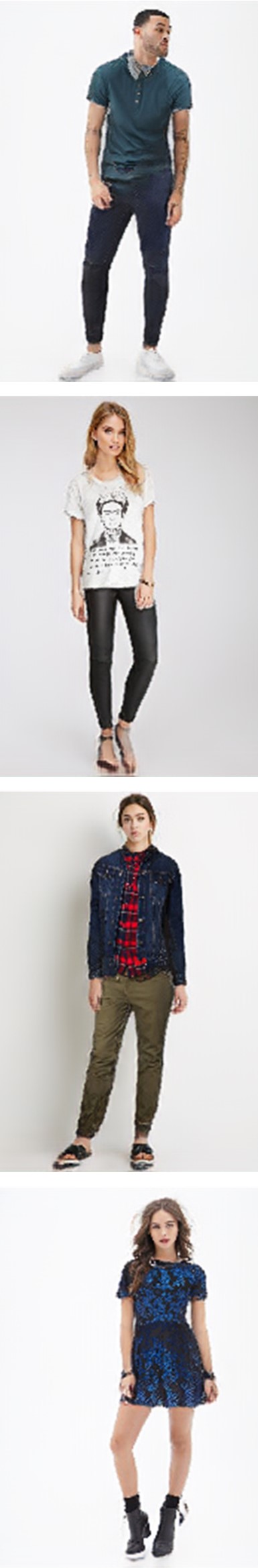} & &
			\includegraphics[width=\linewidth]{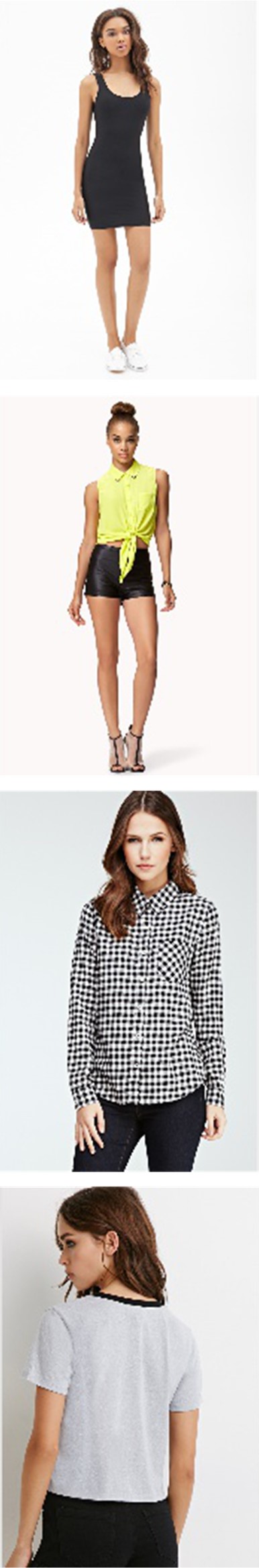} & \includegraphics[width=\linewidth]{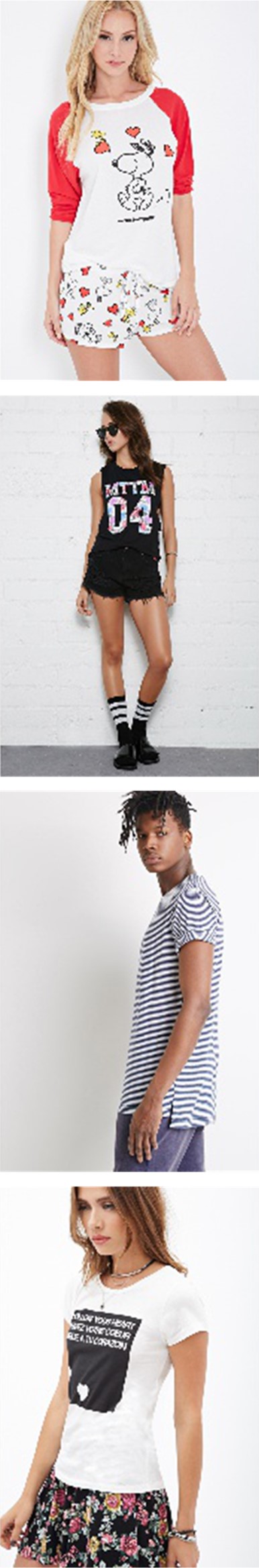} &
			\includegraphics[width=\linewidth]{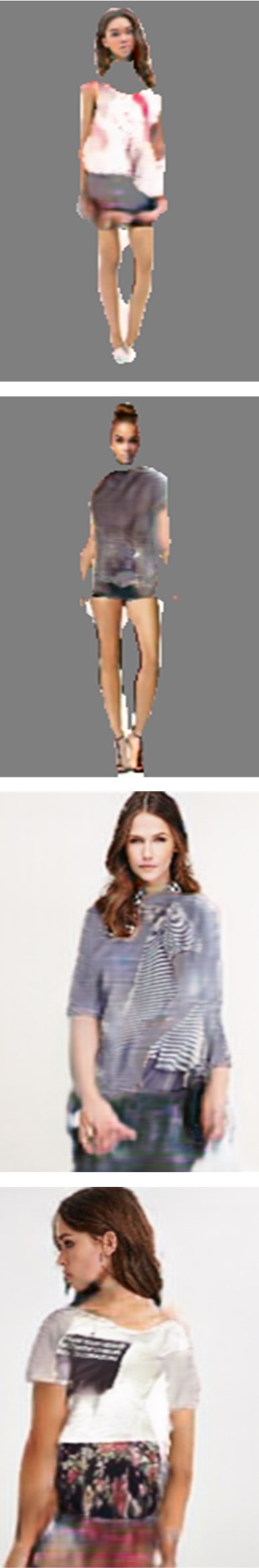} & 
			\includegraphics[width=\linewidth]{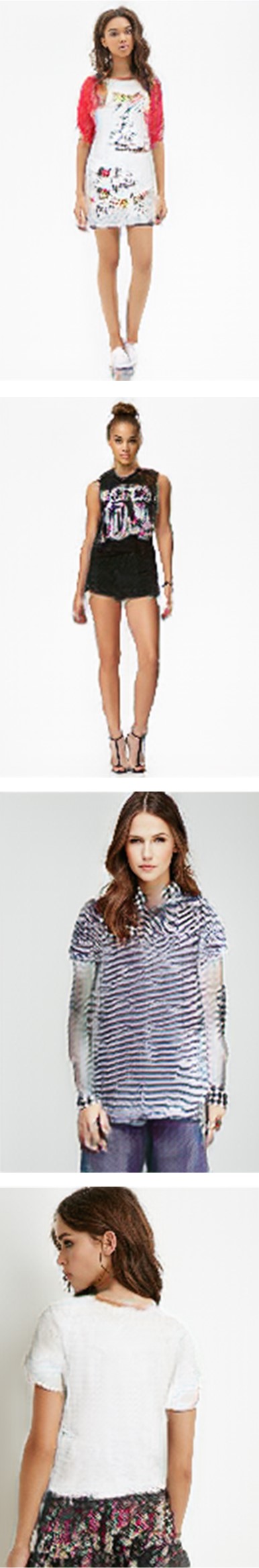} &
			\includegraphics[width=\linewidth]{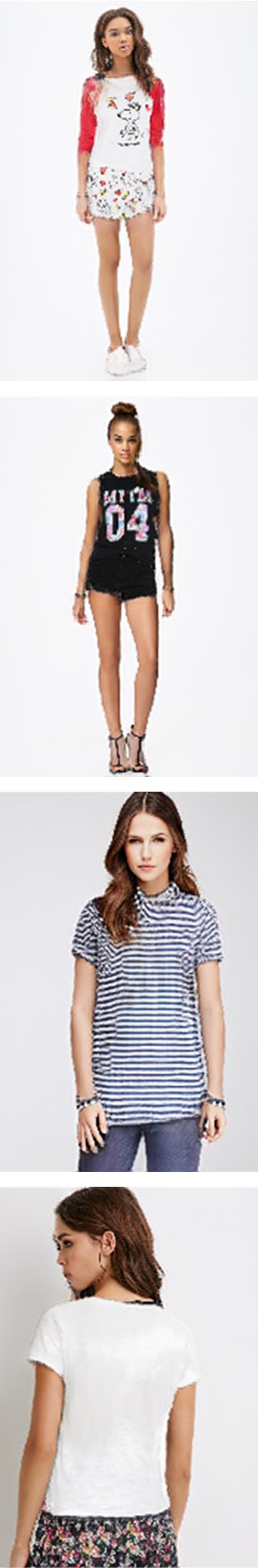} \\
		\end{tabular}
	\end{center}
	\caption{Qualitative comparisons of our method with our baselines.}
	\label{fig:result}
\end{figure*}

	\subsection{Baselines}
	\textbf{ACGPN.} ACGPN is a state-of-art virtual try-on network proposed by Yang \etal~\cite{yang2020towards}, which aims to transfer a stand-alone clothing image onto a reference person. In comparison to previous methods~\cite{han2018viton, wang2018toward}, ACGPN first predicts the target clothing segmentation progressively in two stages, then estimates a TPS warping utilizing STN~\cite{jaderberg2015spatial}. ACGPN shows state-of-art performance on VITON~\cite{han2018viton} dataset with natural deformed clothes and well-preserved non-target body parts.
	
	\textbf{CoCosNet.} CoCosNet stands for the cross-domain correspondence network proposed by Zhang \etal~\cite{zhang2020cross}, aiming to synthesize realistic images according to the examplar images. Different from other methods, CoCosNet establishes a cross-domain correspondence matching to align the examplar image with the target image and then synthesize photo-realistic results, which achieves state-of-art performance in pose-guided human generation task. However, CoCosNet lacks of the ability to generate garment transfer results. To adapt it into our task, we apply the cross-domain correspondence warping to warp the layouts and replace the inputs for the Translation Network in CoCosNet to be the same as ours in the Dynamic Fusion Module. 
	
	To keep the fairness of our experiments, We retrain all aforementioned methods on DeepFashion~\cite{liu2016deepfashion} dataset with the same training set as ours.
	
	\subsection{Qualitative Results}

	Figure~\ref{fig:result} shows the qualitative comparisons of ACGPN, CoCosNet and our model, which indicates that our model synthesizes more convincing results with well-preserved characteristics of clothes and distinct identities of humans. Since ACGPN overlooks the misalignments between the inputing segmentations, it fails to generate the correct target clothing segmentation and estimate reasonable TPS warping, leading to unsatisfying results with messy textures, incorrect body parts and abundant artifacts. Our CoCosNet~\cite{zhang2020cross} baseline eliminates the misalignments by employing the cross-domain correspondence warping. However, cross-domain correspondence warping has a high degree of freedom and fails to preserve the complex clothing patterns, result in visual artifacts such as cluttered textures and blurry boundaries. Benefited from the Layout Prediction Module, our CT-Net adaptively preserves the non-target body parts and generates the occluded parts, leading to realistic garment transfer results with distinct human identities and clear body boundaries. As shown in the first row (right) and second row of the Figure~\ref{fig:result}, the proposed attention mechanism in the Dynamic Fusion Module adaptively combines the advantages of the two complementary warpings and preserves the logos on the desired clothes clearly. In the last row (right), since ACGPN is unaware of the change of the human pose, it wrongly preserves the logo in the back view of the person, while our attention mechanism adaptively drops the logo and synthesizes reasonable back-view image.  

\begin{figure}
		\footnotesize
		\setlength{\tabcolsep}{0.002\textwidth}
		\begin{center}
			\begin{tabular}{*{5}{m{0.085\textwidth}<{\centering}}}
				\includegraphics[width=\linewidth]{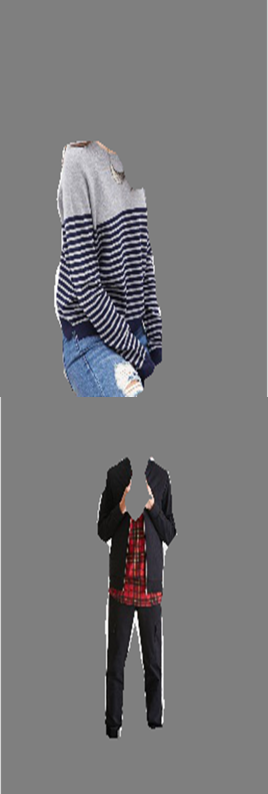} & \includegraphics[width=\linewidth]{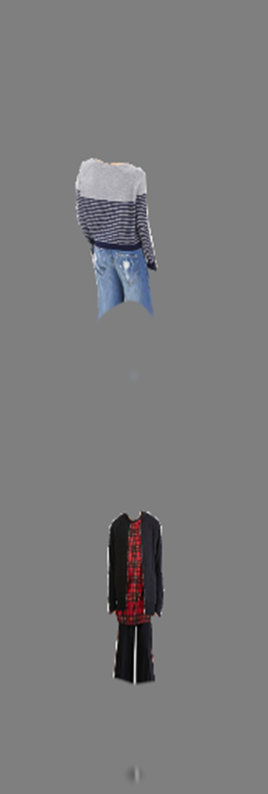} &
				\includegraphics[width=\linewidth]{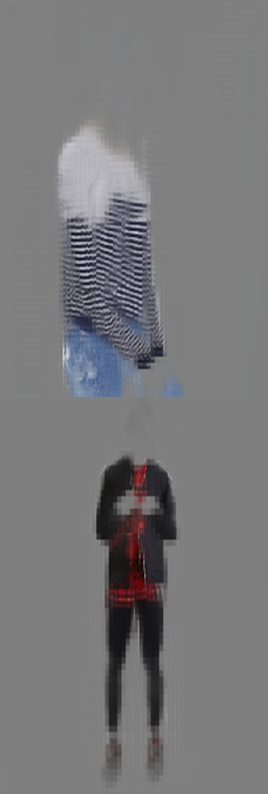} & 
				\includegraphics[width=\linewidth]{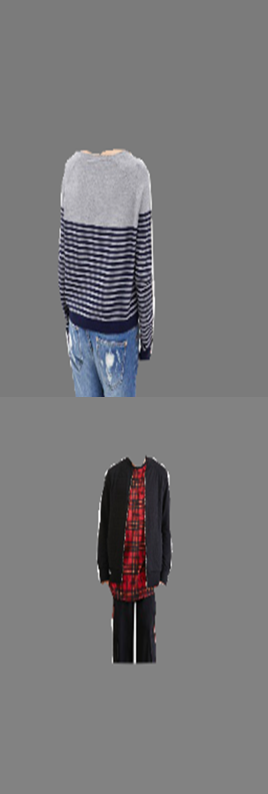} &
				\includegraphics[width=\linewidth]{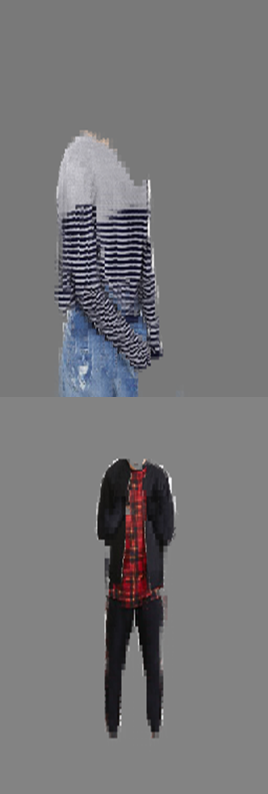} \\
				Target Cloth & ACGPN & CoCosNet & TPS(CWM) & DFW(CWM) \\
			\end{tabular}
		\end{center}
		\caption{Visual comparisons of warping results. DFW(CWM) represents the warping results from DF-guided dense warping estimated in the Complementary Warping Module. }
		\label{fig:warp}
	\end{figure}

	In Figure~\ref{fig:warp}, we visualize warping results from different methods to make further comparisons. Benefited from the joint training of all modules, CT-Net shows superior performance in the estimation of both the DF-guided dense warping and the TPS warping. 
	Please refer to \textit{supplementary materials} for more qualitative results.
	
	\subsection{Quantitative Results}
	We adopt Structural Similarity (SSIM)~\cite{wang2004image} to measure the similarity between the generated images and the real ones. 
	To compute SSIM, we take the same person wearing the same clothes but in different poses as test set. Specifically, inspired by~\cite{siarohin2018deformable, han2019clothflow}, we compute SSIM in three different scales: (\romannumeral1) To isolate the influence from the background, we compute SSIM for human pixels (H-SSIM). (\romannumeral2) To evaluate the accuracy of reconstructed clothes, we compute SSIM for clothing area in the synthesized results (Mask-SSIM). (\romannumeral3) To compare the warping accuracy of different methods, we compute SSIM for warped clothes (Warp-SSIM). For our model, we use the warping results from DF-guided dense warping to calculate Warp-SSIM. Besides, we adopt Inception Score (IS)~\cite{salimans2016improved} to evaluate the quality of our synthesized images. 
	
	Table~\ref{table:quan} reports the quantitative results of our method and baselines. Higher scores are better. As summarized in table~\ref{table:quan}, our method outperforms all baselines by a significant margin. Specially, our model greatly improves the warping accuracy with 0.050 higher Warp-SSIM scores compared to CoCosNet. Moreover, we also achieve higher scores in terms of H-SSIM, Mask-SSIM and IS, which indicates that our method synthesizes more realistic images with well-preserved details.
	
	\begin{table}
		\begin{center}
			\scalebox{0.85}{
			\begin{tabular}{lcccc}
				\hline
				Methods & Warp-SSIM & Mask-SSIM & H-SSIM & IS \\
				\hline\hline
				ACGPN~\cite{yang2020towards} & 0.744 & 0.757 & 0.813 & 3.366 \\
				CoCosNet~\cite{zhang2020cross} & 0.815 & 0.835 & 0.851 & 3.472 \\
				\hline
				w/o PR & 0.857 & 0.913 & 0.919 & 3.495 \\
				w/o LPM & 0.836 & 0.917 & 0.923 & 3.479 \\
				w/o TPS & 0.860 & 0.919 & \textbf{0.931} & \textbf{3.515} \\
				CT-Net (ours) & \textbf{0.865} & \textbf{0.923} & 0.930 & 3.511 \\
				\hline
			\end{tabular}}
		\end{center}
		\caption{Quantitative comparisons of our method with other methods.}
		\label{table:quan}
	\end{table}

	\subsection{Ablation Study} \label{sec:abla}
	We conduct ablation experiments to explore the effectiveness of the main components in our model. In particular, \emph{w/o} PR denotes removing the pose representation $p^M$ inputing to the feature extractor in Complementary Warping Module. \emph{w/o} LPM denotes removing Layout Prediction Module. \emph{w/o} TPS denotes removing the estimation of TPS warping.
	
	\begin{figure}
		\footnotesize
		\setlength{\tabcolsep}{0.002\textwidth}
		\begin{center}
			\begin{tabular}{*{6}{m{0.075\textwidth}<{\centering}}}
				\includegraphics[width=0.075\textwidth]{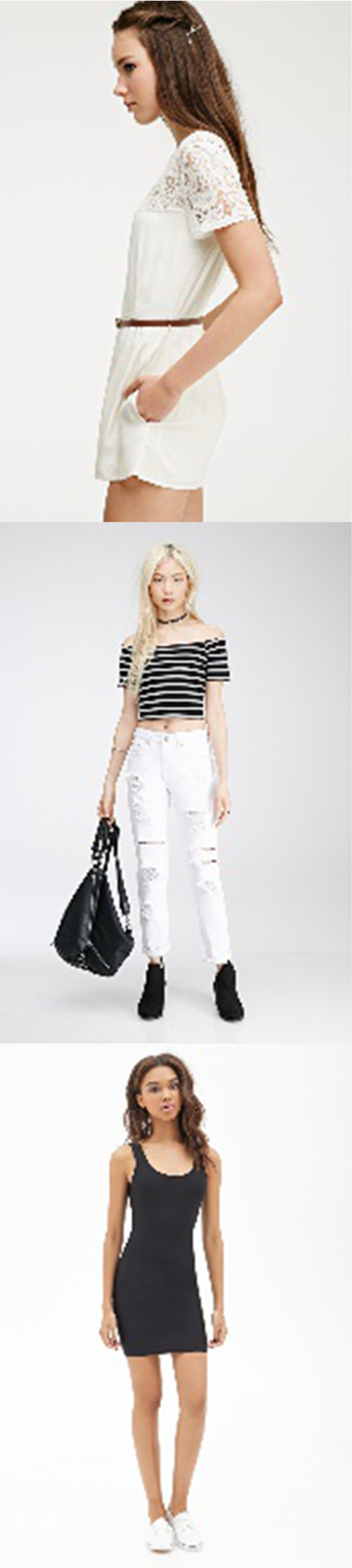} & \includegraphics[width=0.075\textwidth]{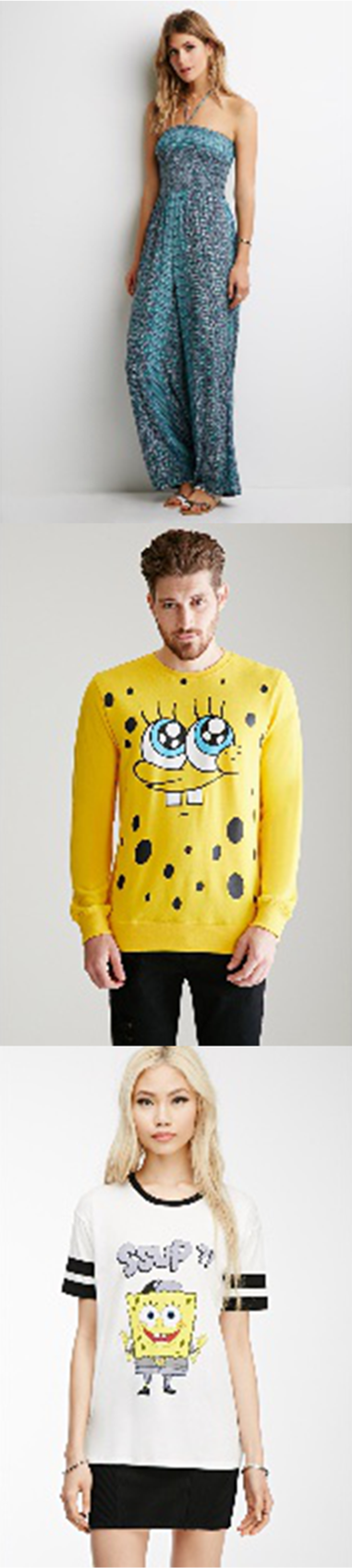} &
				\includegraphics[width=0.075\textwidth]{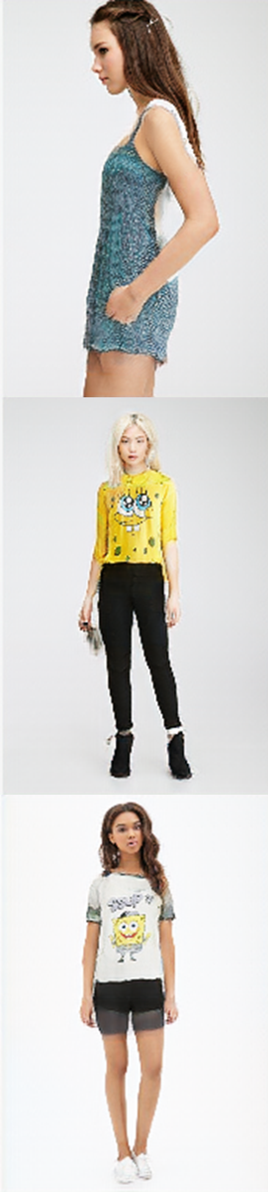} & 
				\includegraphics[width=0.075\textwidth]{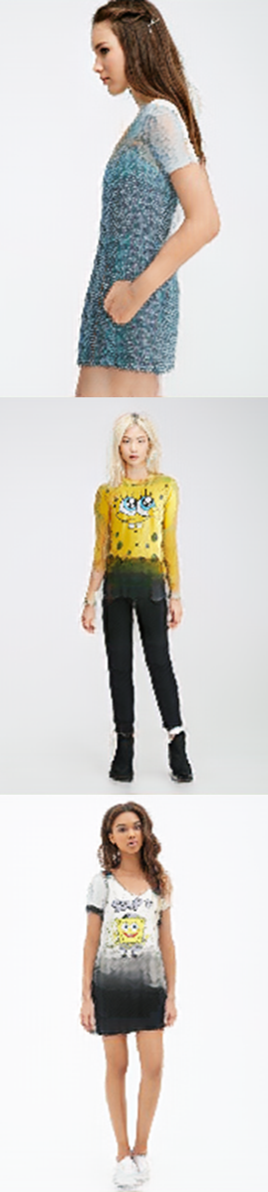} &
				\includegraphics[width=0.075\textwidth]{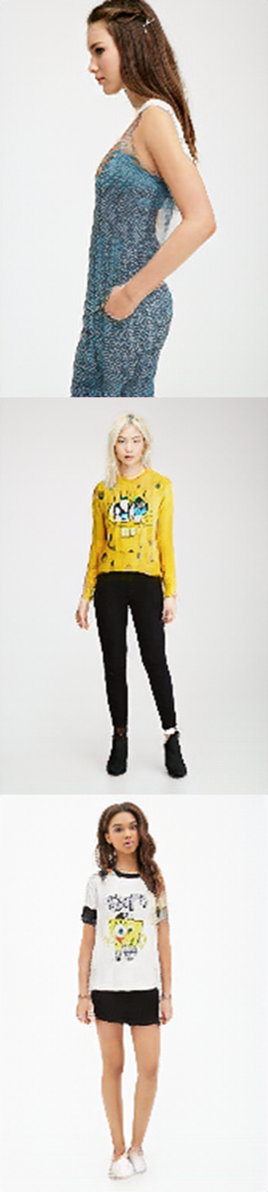} &
				\includegraphics[width=0.075\textwidth]{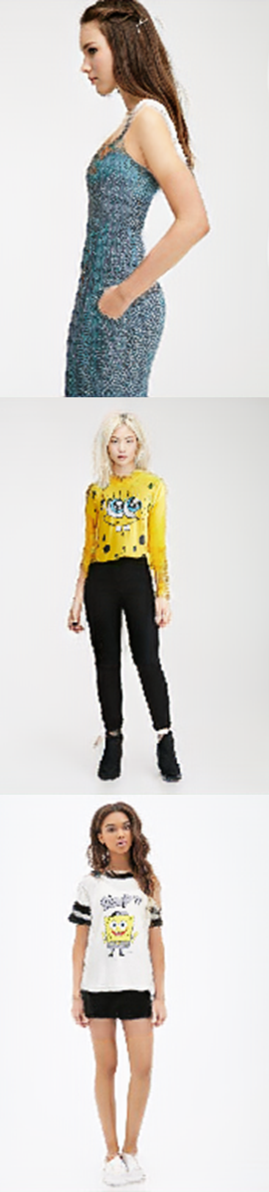} \\
				Target Person & Model Image & \emph{w/o} PR & \emph{w/o} LPM & \emph{w/o} TPS & CT-Net(full) \\
			\end{tabular}
		\end{center}
		\caption{Visual comparisons with ablation methods.}
		\label{fig:ablation}
	\end{figure}

	Table~\ref{table:quan} reports all the results of our ablation experiments. In specific, our full model outperforms all ablation methods by a margin in Warp-SSIM, which indicates that our designs in the network significantly facilitate the estimation of DF-guided dense warping. Combining the advantages of the two complementary warpings, our full model also shows the best performance in reconstructing the clothes and achieves the highest Mask-SSIM scores. Our full model and \emph{w/o} TPS have similar scores in all metrics, since SSIM only roughly measures the local similarity and IS only captures the realism of the images.   
	
	To further demonstrate the superiority of our full model, we visualize some examples to make qualitative comparisons in Figure~\ref{fig:ablation}. Since \emph{w/o} PR and \emph{w/o} LPM can not estimate the warping precisely, artifacts such as incorrect clothing shape (first row) and blurry boundaries (third row) can be observed. Although \emph{w/o} TPS achieves the best scores in terms of H-SSIM and IS, visual results show that our full model synthesizes more photo-realistic images with better-preserved clothing patterns and distinct body parts.

	\section{Conclusion}
	We propose Complementary Transfering Net (CT-Net) for garment transfer with arbitrary geometric changes. In particular, our model adaptively combines two complementary warpings to model different levels of geometric changes and synthesizes photo-realistic garment transfer results with well-preserved characteristics of the clothes and distinct human identities. We introduce three novel modules: \romannumeral1) Complementary Warping Module. \romannumeral2) Layout Prediction Module. \romannumeral3) Dynamic Fusion Module. Experiment results demonstrate that our model significantly outperforms state-of-art methods both qualitatively and quantitatively.

	\section{Acknowledgement}
	This research was conducted in collaboration with SenseTime. This work is supported by A*STAR through the Industry Alignment Fund - Industry Collaboration Projects Grant. This research is supported by the National Research Foundation, Singapore under its AI Singapore Programme (AISG Award No: AISG-RP-2018-003), and the MOE Tier-1 research grants: RG28/18 (S) and RG22/19 (S).

	{\small
		\bibliographystyle{ieee_fullname}
		\bibliography{egbib}
	}
	
\end{document}